\documentclass[11pt]{article}
\usepackage{coling2018}
\usepackage{times}
\usepackage{url}
\usepackage{latexsym}
\usepackage{graphicx}

\title{Event Representation through Semantic Roles: Evaluation of Coverage}

\author{Aliaksandr Huminski \\
  Institute of High Performance Computing \\
  A*STAR, Singapore \\
  {\tt huminskia@}\\
  {\tt ihpc.a-star.edu.sg} \\\And
  Hao Zhang \\
  Artificial Intelligence Initiative \\
  A*STAR, Singapore \\
  {\tt zhang\underline{ }hao@}\\
  {\tt scei.a-star.edu.sg} \\}

\date{}

\begin{document}
\maketitle
\begin{abstract}
Semantic role theory is a widely used approach for event representation. Yet, there are multiple indications that semantic role paradigm is necessary but not sufficient to cover all elements of event structure. We conducted an analysis of semantic role representation for events to provide an empirical evidence of insufficiency. The consequence of that is a hybrid role-scalar approach. The results are considered as preliminary in investigation of semantic roles coverage for event representation. 
\end{abstract}

\section{Introduction}

Almost 50 years ago in his classic article Ch. Fillmore ~\shortcite{fillmore1970} raised a question about the difference in representation of the verbs {\em hit} and {\em break}. Both of them were commonly characterized with [Agent, Patient] role frame. The identical representation was in a conflict with the fact that {\em break} belongs to the verbs of caused change of state while {\em hit} is the verb of impact or surface contact verb. Needless to say, if semantic role theory pretends to be a theory for meaning representation, these two verbs must have different role frames. Not diving deep into the explanation, the solution suggested by Ch. Fillmore was that {\em break} has the frame [Agent, Object] while {\em hit} has the frame [Agent, Place]. So, the issue was resolved inside of semantic role theory by splitting the role of Patient.

This publication had an impact on further investigations which led to a change of a scientific paradigm. First, Beth Levin and Malka Rappaport Hovav ~\shortcite{levin2005,levin2010} paid attention that distinction between {\em hit} and {\em break} can be defined through distinction between manner verbs and results verbs. They pointed out that a study of the English verb lexicon reveals that there can be verbs that describe carrying out activities –- manners of doing; and there can be verbs that describe bringing about results. Manner verbs are {\em hit, stab, scrub, sweep, wipe}, etc. Result verbs are {\em break, clean, crush, destroy, shatter}, etc. Second, manner and result verbs are considered as two parts in a paradigm of event structure ~\cite{Chao}. Event template is represented the following way:
(x ACT{\scriptsize[Manner]} y) CAUSE (y BECOME [State]). Here, manner verbs are defined through the specification of manner (or instrument as a subtype of manner)\footnote{Automatic extraction of manner verbs from WordNet was made by Huminski and Zhang~\shortcite{humi1,humi2}.} located on the left side of event template: 
(x ACT{\scriptsize[Manner]} y), while the result verbs are defined through the specification of resulting state located on the right side:
(x ACT y) CAUSE (y BECOME [State]).

A similar approach comes from cognitive science framework ~\cite{G2017,GW2012,WGW2012} that considers event representation to be based on a 2-vector structure model: a force vector representing the cause of a change and a result vector representing a change in object properties. It is argued that this framework provides a unified account for the multiplicity of linguistic phenomena related to verbs, as well as a cognitive explanation for manner verbs vs. result verbs. Since each vector in the 2-vector structure event model is verb representation, the semantics of a verb reflects only a portion of the event semantics. If the portion contains the force conceptual space underlying actions, it is a manner verb. If the portion contains another conceptual space, it is a result verb. 

\section{Role Frame and Event Structure}

From the point of a role frame, the verbs {\em hit} and {\em break}, as Fillmore concluded, can be distinguished by splitting the role of Patient. 

Although role frame approach makes the meaning representation for them formally different, it doesn't take into consideration that these verbs reflect different sides of event structure.

From the point of event structure (see Table \ref{fig:result0}), 
the verb {\em break} represents unspecified action that causes breaking and well-specified change of state which is the result of the action. In other words, {\em break} doesn't indicate any concrete action and the presence of Agent and Object in its role frame just represents a hypernym of any possible action, something like ACT or AFFECT.
The verb {\em hit} represents well-specified action and unspecified change of state. The role frame [Agent, Place] without any additional means (like selectional restrictions) represents the meaning of {\em hit} partially. For {\em hit} the role frame is suitable but too general since it represents not {\em hit} itself but a class of related actions.\footnote{Interesting, that even for manner verbs a scalar approach was suggested  to provide clear differentiations in meaning. For example, using GL-inspired componential analysis applied to the class {\em run} in VerbNet, six distinct semantic scales -- SPEED, PATH SHAPE,BODILY MANNER, ATTITUDE, ORIENTATION --  emerge ~\cite{Pus2}.}

\begin{table}[t!]
\centering
\begin{center}
\begin{tabular}{|l|c|c|}
\hline   &   specified  &  specified  \\ 
  &  action  &  change of state \\ \hline
verb  {\em hit} & \bf yes & \bf no \\ \hline
verb  {\em break} & \bf no & \bf yes \\
\hline
\end{tabular}
\end{center}
\caption{\label{font-table} Distribution of manner and result verbs inside of event structure.}
\label{fig:result0}
\end{table}

The above-mentioned observations regarding relations between a role frame and event structure trigger the following question. Are semantic roles universal to represent the meaning of manner verbs  and result verbs? In other words, can both manner verbs and result verbs be defined inside of role theory as Fillmore did for {\em hit} and {\em break}?

We are going to formulate a hypothesis as an answer on this question and to provide an empirical evidence for that.

\section{Hypothesis: Role Frames Are Not Sufficient for Event Representation}
By definition, any semantic role is a function of a participant represented by NP, towards an event represented by a verb. If we are going to represent a change of state via roles, we need first to assign a role to {\em state} of a participant, not to a participant itself. Second, a change of state means a change in the  {\em value} of  state in particular direction. Say, the event  {\em heat the water} includes values of state "temperature" for water as a participant. So, to reflect a change of state we need to introduce two new roles: initial value of state and its final value on the scale of increasing values on a dimension of temperature. These 2 new roles look like numbers, not roles, on a scale. It is unclear, what it really means: a role of value.

Nevertheless, semantic role theory was extended beyond the traditional definition of a role in such a way to cover a change of state. There are roles like Attribute, Value, Extent, Asset etc. that match abstract participants, attributes, and their changes. For example, in the sentence {\em Oil soared in price by 10\%}, "price" is in the role of Attribute and "10\%" is in the role of Extent which, according the definition, specify the range or degree of change.

We argue that these attempts contradict the nature of a semantic role. Roles are just one of the parts in event representation that doesn't cover an event completely. While a role is a suitable means for manner verbs representation, a scalar is a suitable means for result verbs.

For instance, the verb {\em kill} has the role frame [Agent, Patient] while the meaning of {\em kill}  contains no information about what Agent really did towards Patient. We represent {\em kill} through an unknown action. Meanwhile, what is important for {\em kill}  is not an action but the resulting change of state: Patient died. And this part of meaning, being not represented at all by roles, can be represented via a scalar change on 1-dimensional scale "alive-dead". The representation of result verbs through a role frame gives us a small piece of their real meaning, since result verbs do not indicate {\em how} it was done but {\em what} was done.

\section{Evidence of Role Frames Insufficiency}

In event template the element BECOME represents a change of state of a participant. Since a change of state means a change in values, the template can be transformed into the following form:

(x ACT{\scriptsize[Manner]} y) CAUSE ([State{\scriptsize 1}]{\scriptsize y} $\,\to\,$[State{\scriptsize 2}]{\scriptsize y})
\\
where [State{\scriptsize i}]{\scriptsize y} is a value {\em i} of state of {\em y}. 

In this case, since event template contains a pair of participants from the both sides, we expect to get a mutual dependency for roles of x and y (something like Agent-Theme), representing the left side, and a mutual dependency for roles of [State{\scriptsize 1}]{\scriptsize y}  and [State{\scriptsize 2}]{\scriptsize y} (something like Origin-Destination), representing the right side.\footnote{In FrameNet there are explicit relations "Requires" and "Excludes" that define kind of dependency between roles.}

We will verify this assumption. But before verification a role set resource should be taken.

\subsection{Role set resource}

Semantic role theory is one of the oldest constructs in linguistics. As a result of that, variety of resources with different 
sets of semantic roles has been proposed. There are 3 types of resources depending on the level of role set granularity. 
The first level is very specific with roles like "eater" for the verb {\em eat} or "hitter" for the verb {\em hit}. The third level is very general with only two “proto-roles” ~\cite{dowty}.
The second level is located between them and contains, to the best of our knowledge, from 10 to 50 roles approximately.

We made a decision to start our investigation from Verbnet (VN) that contains 30 roles and belongs to the second level of role set recourses. It was chosen because of the following 4 reasons. First, VN is the largest domain-independent computational verb lexicon currently available for English ~\cite{schuler2005}. Second, it provides semantic role representation for all verbs from the lexicon. Third, the roles are not so fine-grained as in FrameNet ~\cite{fillmore2002} and not so coarse-grained as in Propbank ~\cite{palmer2005}. Fourth, VN was considered together with the LIRICS role set for the ISO standard 24617-4 for Semantic Role Annotation ~\cite{Pet,Claire,bunt2013conceptual}.

In this paper, we use the latest VerbNet version 3.2b that contains 6394 verbs.

\subsection{Extraction of Role Frames}

To analyze the relations across roles, we need first to extract the verbs with corresponding role frames from VN. The verbs are organized through 498 verb classes: 277 root verb classes (RVC) and 221 sub-root verb classes (sub-RVC). Each RVC and sub-RVC has its own role frame. For example, RVC {\em destroy-44} has the role frame [Agent, Patient, Instrument]. It is a common situation when sub-RVC inherits roles from RVC and has the same role frame. In this case, sub-RVC is merged with their parent RVC. We found only 13 sub-RVCs that have their own additional roles besides the roles inherited from parent RVCs.

To access VN, JVerbNet toolkit 1.2.0. is used ~\cite{jverbnet2012}. To extract all RVCs and sub-RVCs that are different from their parent RVCs, we use Depth-First-Search method ~\cite{tarjan1972dfs} that allows making recursive search from RVC to its sub-RVCs.

As a result, 498 verb classes were compressed to 290 verb classes (277 RVCs and 13 sub-RVCs). For these 290 classes we extracted the name of the class, the number of its members and the roles in the class frame. 
Example: class {\em break-45.1}; members: 24; role frame: [Agent, Patient, Instrument, Result].

\subsection{Roles as Vectors}
Since there are 30 different roles in VerbNet, we constructed a 30-dimensional vector for each extracted verb class. Each element of the vector represents one role: if a verb class has a role in its frame, the corresponding position is set as 1, otherwise 0.
If each verb class is represented by 30-dimensional vector, each role is represented by 290-dimensional vector according the number of classes.

The last step is to represent a role not as a vector based on classes but as a vector based on verbs. A class contains a group of verbs that share the same role frame. So, to expand a vector on verbs we just need to expand 1s or 0s n times, where n is the number of members of a class. Using this way, each role is transformed from 290-dimensional vector into 6394-dimensional vector.

\subsection{Roles: Mutual Dependency vs. 1-way Dependency}

The method for calculation of dependency or contextual occurrence includes two steps:

\begin{enumerate}
\item For a pair of roles, such as "Agent-Theme", we get their vectors and count the number of positions with "1" for both vectors: $val\textsubscript{$common$}$.
\item After computing $val\textsubscript{$common$}$, we count all "1" for the first vector (Agent): $sum\textsubscript{$agent$}$, and all "1" for the second vector (Theme): $sum\textsubscript{$theme$}$. Then we compute 2 ratios: 

$P(Theme|Agent) = \frac{val _{common}}{sum _{agent}}\times100$ 

$P(Agent|Theme) = \frac{val _{common}}{sum _{theme}}\times100$.
\end{enumerate}

\begin{figure*}
\centering
\includegraphics[width=1\linewidth]{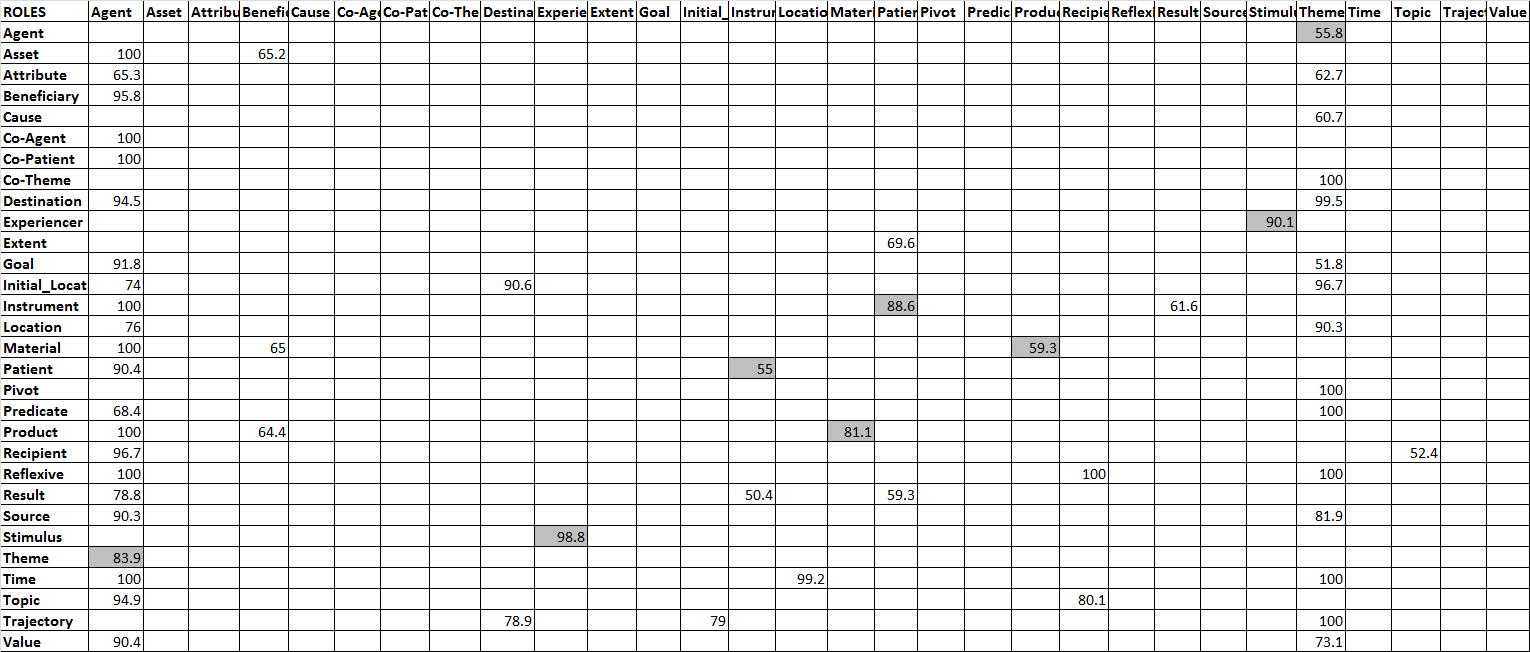}
  \caption{Visualization of Roles Dependency.}
  \label{fig:result}
\end{figure*}

We apply this method to all the role pairs and the results are shown in the Figure \ref{fig:result} with the threshold of 55\% of dependency. 

One can see that the behavior of roles is different. All roles can split on 2 groups: mutually dependent and mutually independent. The following pairs are mutually dependent 8 roles (highlighted in grey):

Agent (55.8) -- Theme (83.9)

Experiencer (90.1) -- Stimulus (98.8)

Instrument (88.6) -- Patient (55)

Material (59.3) -- Product (81.1)

Among them, six roles (except the pair "Material -- Product" which is a non-scalar change) belong to the left side of event structure for manner verbs representation. It means, 22 roles left are one-way dependent (see white cells with numbers). They don't form pairs for result verbs representation (right side of event structure). It is an evidence of roles insufficiency. 

\section{Conclusion and Future Work}
\label{sec:length}

Based on Verbnet as a role set resource we described an empirical evidence of role frames insufficiency that supports the hypothesis that role frames as a tool for meaning representation don't cover event structure completely. 
As a consequence of that, another paradigm -- scalar approach -- is needed to fill up the gap. The hybrid role-scalar approach looks promising for meaning representation and will be elaborated in future for event structures.
We do not examine yet to what extend our argumentation also applies to other resources (e.g. FrameNet). That is why the results are considered as preliminary ones.


\end{document}